# Road User Classification from High-Frequency GNSS Data Using Distributed Edge Intelligence

Lennart Köpper, Thomas Wieland

*Abstract*— **Real-world traffic involves diverse road users, ranging from pedestrians to heavy trucks, necessitating effective road user classification for various applications within Intelligent Transport Systems (ITS). Traditional approaches often rely on intrusive and/or expensive external hardware sensors. These systems typically have limited spatial coverage. In response to these limitations, this work aims to investigate an unintrusive and cost-effective alternative for road user classification by using high-frequency (1-2 Hz) positional sequences. A cutting-edge solution could involve leveraging positioning data from 5G networks. However, this feature is currently only proposed in the 3GPP standard and has not yet been implemented for outdoor applications by 5G equipment vendors. Therefore, our approach relies on positional data, that is recorded under real-world conditions using Global Navigation Satellite Systems (GNSS) and processed on distributed edge devices. As a starting point, four types of road users are distinguished: pedestrians, cyclists, motorcycles, and passenger cars. While earlier approaches used classical statistical methods, we propose Long Short-Term Memory (LSTM) recurrent neural networks (RNNs) as the preferred classification method, as they represent state-of-the-art in processing sequential data. An RNN architecture for road user classification, based on selected motion characteristics derived from raw positional sequences is presented and the influence of sequence length on classification quality is examined. The results of the work show that RNNs are capable of efficiently classifying road users on distributed devices, and can particularly differentiate between types of motorized vehicles, based on two- to four-minute sequences.**

## I. INTRODUCTION

Real-world traffic consists of various types of participants ranging from vulnerable road users, such as pedestrians and cyclists, to light vehicles, like motorcycles and small passenger cars, and up to heavy trucks. Determining the *type* of these participants – commonly referred to as vehicle classification, or more broadly, *road user classification* – has been a key challenge since the early 2000s across various applications in Intelligent Transport Systems (ITS), including traffic management and monitoring, highway tolling, traffic flow estimation, and urban planning [1], [3], [4]. Of course, the categorization of road user types strongly depends on the context and the specific application. Several definitions can be considered, ranging from broad categories, like those mentioned above (pedestrians, cyclists, …), to arbitrarily fine-grained categorizations. For example, a well-known definition considering vehicle classes is the categorization by the United States Federal Highway Administration that proposes 13 categories based on the vehicle weight, length, number of axles and axle spacing [5].

Traditional approaches use various types of hardware sensors, chosen according to the application's requirements and the level of classification detail needed. Some applications require the installation of physical components on road or pavement segments. These fixed-location sensor systems include pneumatic tubes, inductive loops, piezoelectric sensors, and weigh-in-motion systems. Approaches utilizing such sensor systems are generally categorized as traffic-intrusive methods since they frequently require on-site work that interferes with traffic [4], [6], [7]. In exchange some of these methods can achieve high classification performance, even providing a full 13-class vehicle classification with high accuracy. However, the high installation and maintenance costs of such systems limit their widespread deployment [4], [5], [8]. Furthermore, literature highlights that intrusive methods are restricted to certain applications and prone to errors or limitations under specific conditions. For example, pneumatic tubes do not perform well on high-speed roads and inductive loop detector-based systems encounter issues in congestion situations [1], [4], [5], [8].

In response to these limitations, less intrusive alternatives have gained increasing popularity, including fixed-position and mobile sensor systems, as well as vehicle-mounted sensors that utilize infrared, ultrasonic, and radar technologies. Those systems offer a trade-off in costs and applicability on one hand, and granularity on the other. Moreover, such sensor systems may be affected by several environmental and traffic conditions [1, 5]. Another yet unmentioned unintrusive approach is the classification based on images or videos obtained from (surveillance) cameras [4], [10], [11]. Advances in computer vision and deep learning techniques, particularly convolutional neural networks (CNNs), have significantly improved the performance of camera-based classification systems. Given sufficient training data for the underlying networks, modern camera-based approaches can classify vehicles with high granularity, even under adverse conditions such as noisy or dark images [12], [13], [14]. However, they still face certain limitations, e.g., when the line of sight between the camera and the vehicle is obstructed by

S Submitted: 30 August 2024. This work was supported in part by the German Federal Ministry of Digital Affairs and Traffic under grant 165GU102C. Corresponding author: Thomas Wieland.

Authors' Address: Coburg University of Applied Sciences and Arts, Faculty of Electrical Engineering and Computer Science, 96450 Coburg, Germany (e-mail: thomas.wieland@hs-coburg.de).

environmental factors.

To overcome the limitations of existing approaches, a cutting-edge solution could involve the use of 5G networks, which are generally capable of determining positional data for each user device in a certain cell [15], [16]. During the establishment of our 5G campus networks, we planned to utilize this data for the classification task. However, when our project started in 2021, the positioning features were not implemented in the 5G core network, and later, support was limited to indoor scenarios. As a preliminary substitute, Global Navigation Satellite System (GNSS) data was chosen. The widespread adoption of GNSS trackers in general-purpose devices, particularly smartphones, facilitates comprehensive spatiotemporal trajectory data collection. Typically used for navigation or geo-localization purposes, such devices can continuously measure their location with a high sampling rate of around one sample per second. GNSS data can be easily processed to further obtain speeds, accelerations, and decelerations. The frequency, quality, and information content of GNSS data generated from modern smartphones combined with the fact that using GNSS infrastructure does not incur extra costs (since the system is already designed and deployed) therefore motivates the use of this data within several transportation-related applications [5], [12], [19]. However, the utilization of general-purpose GNSS data for such applications is still relatively unexplored in literature, with some works in driving behavior analysis [20], transportation mode prediction [20], [21], [22] and just a few approaches for vehicle classification [5], [6], [12].

Simoncini et al. [5] (2018) investigated vehicle classification into small-, medium-, and heavy-duty categories using low-frequency GPS trajectories with a mean sampling interval of 90 seconds. The extensive dataset contained about one million trajectories, each with at least 20 timesteps, corresponding to a duration of approximately 30 minutes. Features such as distance, time, velocity, and road type were extracted for each GPS sample pair. Recurrent neural networks (RNNs) with Long Short-Term Memory (LSTM) layers were employed for classification. The best model achieved a Macro-Recall of 0.79 for binary classification of light- and heavy-duty vehicles, but only 0.66 for all three classes due to difficulties in distinguishing medium- from small-duty vehicles.

Dabiri et al. [12] (2020) used a Convolutional Neural Network (CNN) architecture for a similar task. Their dataset included about 230,000 trajectories with an average sampling interval of 17 seconds and a duration of approximately 20 minutes. The trajectories were preprocessed to capture vehicle motion characteristics and roadway features. The CNN model achieved a macro-recall of 0.74 for binary classification but struggled with all three classes, resulting in a macro-recall of 0.63.

The authors of both [5] and [12] suggest that higher-frequency GNSS data might improve their models' ability to capture vehicle motion variations, potentially leading to more granular and reliable classifications with shorter observation durations.

More recent approaches either rely on camera infrastructure [23], [24], [11], on vehicular ad-hoc networks (VANETs) [8] or road sensors [7].

This paper investigates road user classification using GNSS data by evaluating a deep learning approach to classify high-frequency positional sequences (trajectories) recorded in real-world conditions with general-purpose devices. As a starting point, we focused the classification task on scenarios within medium-sized towns and rural communities, limiting it to four types of road users, namely *pedestrians*, *cyclists*, *motorcycles,* and *passenger cars*. The data was specifically collected for this paper utilizing a smartphone app developed for this project. After training, the model can be deployed on smartphones, enabling classification on mobile devices acting as distributed edge devices.

We chose LSTM-RNNs for classification as they represent the current state-of-the-art in machine learning for processing sequential data. An RNN architecture for road user classification based on selected motion characteristics is presented and the influence of sequence length on classification quality is examined. We generate multiple datasets and preprocess them to derive suitable, generalizable features. The results of the work show that RNNs are highly capable of reliably classifying volatile non-motorized road users, such as pedestrians or cyclists, based on even short sequences. The classification of motorized vehicles is considerably more challenging. However, the capability to efficiently classify corresponding sequences improves with increasing sequence lengths. Our method achieves an F1-score of 0.82 for passenger cars and 0.76 for motorcyclists based on sequences of just two-minute length or of 0.85 and 0.77 based on four-minute sequences respectively. It therefore demonstrates its potential to classify road users (especially motorized vehicles) not only reliably but also with shorter observation periods than existing approaches that utilize lower-frequency GNSS data.

The remaining sections are organized as follows: Section II presents the methodology we followed, describing the utilized data and its preprocessing, as well as the generalized recurrent neural network architecture we used in our experiments. The results of these experiments are subsequently presented in Section III, followed by our conclusion and remarks on future research in Section IV.

## II. METHODOLOGY

In this section, we outline the methodology of our approach, describing data acquisition and preprocessing as well as the utilized neural network architecture.

### A. Data Acquisition

A primary challenge in GNSS-based road user classification is the limited availability of high-quality, publicly accessible datasets, particularly for vulnerable and critical user groups, such as cyclists and motorcyclists.

The dataset used in this work was therefore collected by a relatively small group of just six volunteers in everyday traffic, using a mobile app specifically developed for this purpose. A total of 165 real-world road user trajectories were collected, representing a duration of approximately 32 hours. The trajectories collected by the app are labeled and can be categorized in four classes. About 34.6% of the collected data represent passenger cars, 24.5% cyclists, 20.3% motorcyclists and 20.6% pedestrians. Each trajectory comprises a sequence of timestamps, latitudes, longitudes, and accuracy measurements recorded with a sampling frequency of approximately one second. All trajectories were collected in the Upper Franconia region of Germany, where Coburg University is located. Consequently, the data primarily originates from traffic in medium-sized towns and rural areas, with no representation of urban traffic.

To investigate the impact of different sequence lengths, we first generated three datasets from the collection of trajectories by truncating each trajectory to multiple sequences with lengths of 60 (1min), 120 (2min) or 240 (4min) data points, respectively. This truncation method also provides the benefit of increasing our sample size as a side effect. Sequences shorter than the expected length are discarded at this stage. The remaining sequences are split into a training set (75%) and a test set (25%) using stratified sampling to preserve class ratios. The training set is further divided into actual training data (80%) and validation data (20%). This split is also stratified.

We also created datasets with two-second sampling intervals by removing alternate data points in each sequence, yielding three additional datasets with a length of 30 (1min), 60 (2min) or 120 (4min) data points respectively. The rationale behind increasing the sampling interval is to introduce greater temporal and spatial distances between the recorded positions, potentially reducing noise in the sequences and its impact on the features calculated in the next subsection.

*B. Feature Extraction*

Let $\{\boldsymbol{p}_t\}_{t=1}^T = \{\boldsymbol{p}_1, \ldots, \boldsymbol{p}_T\}$ be a sequence of $T$ datapoints (GNSS-samples) that represents a single trajectory in a given dataset. Each datapoint $\boldsymbol{p}_t = [\text{lat}_t, \text{lon}_t, t_t, \text{acc}_t]$ is a vector comprising a timestamp $t_t$ in milliseconds (ms) since epoch, the measured latitude $\text{lat}_t$ and longitude $\text{lon}_t$ in decimal degrees, as well as the estimated accuracy $\text{acc}_t$.

To obtain the input sequences for the recurrent neural network, several transformations were applied to the given sequences. To reduce the strong bias arising from the geographical and temporal constraints of the collected data, as well as the limited number of data collectors, each sequence $\{\boldsymbol{p}_t\}_{t=1}^T$ is transformed to an input sequence $\{\boldsymbol{x}_t\}_{t=1}^T$ with each input-vector $\boldsymbol{x}_t$ consisting of the following features:

- *Time difference*: the time elapsed between consecutive measurements in seconds:
$$\Delta t_t = \frac{t_t - t_{t-1}}{1000}. \quad (1)$$
For $t = 1$, we set $\Delta t_1$ to 0.
- *Velocity*: the interval velocity, representing the rate at which a road user covered the distance between consecutive measurements, in meters per second:
$$v_t = \frac{s_t}{\Delta t_t}. \quad (2)$$
Here $s_t$ is the great-circle (spherical) distance, which can be calculated by the *Haversine formula*. For $t = 1$, $v_1$ is later set to the value of $v_2$ (*backward-filling*).
- *Acceleration*: the positive interval acceleration, indicating the increase in velocity of a road user in the current interval compared to the last interval, in meters per square second:
$$a_t^+ = \max\left(\frac{v_t - v_{t-1}}{(\Delta t_t + \Delta t_{t-1})/2}, 0\right). \quad (3)$$
For $t \leq 2$, we set $a_t^+$ to 0.
- *Deceleration*: the negative interval acceleration (deceleration), indicating the decrease in velocity of a road user in the current interval compared to the last interval, in meters per square second:
$$a_t^- = -\min\left(\frac{v_t - v_{t-1}}{(\Delta t_t + \Delta t_{t-1})/2}, 0\right). \quad (4)$$
For $t \leq 2$, we set $a_t^-$ to 0.
- *Bearing rate*: The absolute rate at which a road user has changed its direction of movement (bearing/heading) since the last interval in radians per second:
$$\omega_t = \frac{\pi - ||\beta_t - \beta_{t-1}| - \pi|}{(\Delta t_t + \Delta t_{t-1})/2}. \quad (5)$$
Here $\beta_t$ is the bearing between consecutive measurements, in radians ($0 \triangleq \text{north}, \pi/2 \triangleq \text{east}, -\pi/2 \triangleq \text{west}, \pi \triangleq \text{south}$). The bearing between two coordinates on a sphere, can be calculated by the following formula:
$$\beta_t = \text{atan2}(x, y) \in (-\pi, \pi] \quad (6)$$
with
$$x = \cos(\widetilde{\text{lat}}_t) \times \sin(\widetilde{\text{lon}}_t - \widetilde{\text{lon}}_{t-1}) \quad (7)$$
and
$$y = \cos(\widetilde{\text{lat}}_{t-1}) \times \sin(\widetilde{\text{lat}}_t) - \sin(\widetilde{\text{lat}}_{t-1}) \times \cos(\widetilde{\text{lat}}_t) \times \cos(\widetilde{\text{lon}}_t - \widetilde{\text{lon}}_{t-1}). \quad (8)$$
Here $\sim$ indicates, that the coordinates have been converted to radians. For $t \leq 2$ $\beta_t$ is set to 0.

Note that if duplicate positions occur in the sequences, this may lead to some $\Delta t_t$ values being equal to 0, which prevents the correct calculation of $v_t$ and $\beta_t$. Therefore, the features $a_t^+, a_t^-$ and $\omega_t$, which depend on these, cannot be correctly computed either. To tackle this problem, non-computable $v_t$ and $\beta_t$ are directly set to their last computable values within the sequence (*forward-filling*). An advantage of this approach is that the dependent features can be calculated based on the filled values, albeit with a certain degree of error.

Finally, the preprocessing of the data is completed by standardizing the calculated features.

*C. Neural Network Architecture*

The goal of this work is to classify each GNSS-trajectory given in the datasets mentioned in section II.A according to the type of road user that generated it. Therefore, each trajectory was transformed to a sequence of input vectors $\{\boldsymbol{x}_t\}_{t=1}^T$, $\boldsymbol{x}_t = [\Delta t_t, v_t, a_t^+, a_t^-, \omega_t] \in \mathbb{R}^5$, computed according to the formulae in section II.B. Given that the classification task involves sequential data, we opted to use recurrent neural networks, particularly Long Short-Term Memory networks, for our experiments due to their effectiveness in capturing temporal dependencies. Unlike traditional feedforward architectures or convolutional neural networks (CNNs), which are optimized for static or spatial data respectively, RNNs are inherently suited for processing sequential information [25], [26]. While we acknowledge that other architectures such as transformers [27] have shown strong performance in various domains of sequential-processing, we believe that RNNs, offer a simpler, more lightweight architecture suitable for deployment on edge devices with relatively limited computational resources. Moreover, RNNs with LSTM cells are advantageous in this scenario because they are specifically designed for sequential data and can capture the temporal and dynamic nature of positional data. Other alternatives like bilinear neural networks [28], while potentially useful for specific relationships in static data, but typically requiring fixed-size inputs, making them less flexible for variable-length sequences in positional data, are not inherently suitable for modeling time-series data or trajectories where past and future positions are crucial for classification.

Since it is difficult to estimate the network's optimal architecture and hyperparameters, a generalized RNN architecture, as shown in Fig. 1, is adopted and later used to consider different network configurations. According to the depiction, the RNNs considered consist of an input and output layer as well as three separate groups of hidden layers. The input layer is fixed to contain exactly five fully connected input nodes, one for each input feature listed above. The output layer consists of exactly four fully connected neurons activated by the Softmax function, each representing one of the possible road user types. By utilizing the Softmax function the value of an individual output neuron can be interpreted as the probability that a sequence, based on every processed timestep up to $t$, belongs to the class represented by the respective neuron.

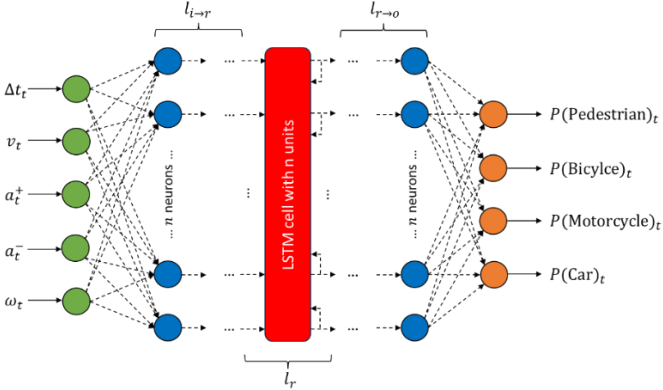

**Fig. 1.** Utilized RNN architecture

Regarding the hidden layers, the generalized architecture provides high flexibility for adapting the network, e.g., through hyperparameter tuning. Nevertheless, some structure is provided by the arrangement of the hidden layers into three separate groups. This arrangement is inspired by the RNN architecture proposed by Simoncini et al. in [5]. Accordingly, the input layer is followed by a group of $l_{i \to r}$ input-to-recurrent fully connected feed-forward layers, $l_r$ recurrent LSTM cells and $l_{r \to o}$ recurrent-to-output fully connected feed-forward layers, finally leading to the output layer. The number of layers in each group ($l_{i \to r}, l_r, l_{r \to o}$), the number of neurons/units per hidden layer/cell $n$ and the activation function $\varphi(\cdot)$ are hyperparameters of the generalized architecture. $n$ and $\varphi(\cdot)$ are equally set for every hidden layer, to mitigate the complexity of the hyperparameter tuning process.

The idea behind arranging the hidden layers into three groups is as follows: The input-to-recurrent group operates independently on each temporal input. This means it always applies the same (learned) function to each input vector in what can be thought of as a feature extraction phase. In contrast, the recurrent group operates sequence-wise, meaning that it applies a function that depends not only on the current temporal input, but also on the previous state of the group. Since this state depends on every previous input, one can think of this group as the memory of the network, which is meant to decode and store temporally encoded information. The final output of the recurrent group is a reduction of the whole input sequence to a single fixed-length vector, which is subsequently post-processed by the recurrent-to-output group to finally produce the output of the network [5]. It is important to note that networks with the given architecture in fact generate one prediction for each temporal input and can therefore be classified as a sequence-to-sequence RNN. However, with respect to the given classification task, one could consider the network as a sequence-to-vector RNN because its final output provides the most likely prediction. Nevertheless, the sequence-to-sequence design will come in handy when training the networks, because this allows the computation of gradients not solely with respect to the loss of the final output but with respect to the total loss across every temporal output. This increases the number of gradients backpropagated, resulting in an acceleration and stabilization of the training process.

Each temporal output of the RNNs represents a probability vector, with values ranging from zero to one and summing up to a total of one. Therefore, *categorical cross-entropy* is used as the loss function for training optimization. To further accelerate the training, the *Adam optimization* method is used. To keep the hyperparameter tuning process simple, we use the default parameters proposed in [29]. The training is performed with minibatches of size 32. To prevent overfitting, we implement *early stopping* with a patience period of 10 epochs. Training halts if there is no further decrease in loss on the validation set during this period. Afterwards, the weights from the epoch with the best loss are restored.

The Glorot (Xavier) uniform initialization is used to initialize the model weights. This method has proven effective in mitigating vanishing gradient issues due to neuron saturation during training [30]. To identify optimal hyperparameter configurations, the grid search method is utilized to tune the hyperparameter of the generalized RNN architecture with respect to a specific dataset.

TABLE I
STATIC PARAMETERS USED IN RNN TRAINING

| Loss Function | Categorical Cross Entropy |
|---|---|
| Optimizer | Adam Optimization |
| Adam Learning Rate ($\eta$) | 1e$^{-3}$ |
| Adam $\beta_1$ | 0.9 |
| Adam $\beta_2$ | 0.999 |
| Weight Initialization | Glorot / Xavier Uniform Initialization |
| Minibatch Size | 32 |
| Early Stopping Patience | 10 Epochs |

The grid search is performed along the lists of parameter values mentioned in Table II. These lists collectively form a grid, comprising a total of 216 possible hyperparameter combinations. For each of these combinations a RNN model is constructed according to the specified hyperparameters, trained using the training data and subsequently validated by assessing its loss on the validation data. The best hyperparameters can be easily identified as the specific combination yielding the model with the lowest loss across the explored grid.

TABLE II
PARAMETER GRID FOR GRID SEARCH
HYPERPARAMETER TUNING

| Input-to-Recurrent Layers ($l_{i \to r}$) | 1, 2, 4 |
|---|---|
| LSTM Cells ($l_r$) | 1, 2, 4 |
| Recurrent-to-Output Layers ($l_{r \to o}$) | 1, 2, 4 |
| Number of Neurons / Units per Layer / Cell ($n$) | 32, 64, 128, 256 |
| Activation Function of Hidden Layers ($\varphi(\cdot)$) | tanh, ReLU |

III. EXPERIMENTAL RESULTS

In this section, we present the experimental results of the hyperparameter optimization as well as the evaluation process and provide an interpretation of the final models.

*A. Results of Hyperparameter Tuning*

As already mentioned in subsection II.A, we consider a total of six variants of our dataset for the road user classification task. These variants differ in terms of the sampling interval and the length of the corresponding sequences. Both variables can be thought of as additional hyperparameters for this specific classification task.

For each dataset we performed a grid search, as described in subsection II.C, for finding good hyperparameters according to our RNN architecture proposed there. Table III lists the result of this grid search: the hyperparameters yielding the best found RNN model for the specific datasets.

TABLE III
RESULTS OF GRID SEARCH

| Sampling interval | Seq. Length | $l_{i \to r}$ | $l_r$ | $l_{r \to o}$ | $n$ | Valid. Loss |
|---|---|---|---|---|---|---|
| 1 s | 60 (1 min) | 1 | 1 | 2 | 128 | 0.4562 |
| 1 s | 120 (2 min) | 4 | 1 | 1 | 64 | 0.4411 |
| 1 s | 240 (4 min) | 4 | 2 | 1 | 128 | 0.4136 |
| **2 s** | **30 (1 min)** | **4** | **1** | **2** | **64** | **0.4534** |
| **2 s** | **60 (2 min)** | **4** | **1** | **2** | **64** | **0.4282** |
| **2 s** | **120 (4 min)** | **2** | **1** | **2** | **128** | **0.3967** |

Regarding the found hyperparameters, some notable observations can be made.

- Within this specific classification task, the hyperbolic tangent seems to outperform the ReLU function as activation function $\varphi(\cdot)$ for the hidden layers, consistently achieving a lower validation loss.
- The increase of the sampling interval to two seconds results in a reduction of the validation loss across all sequence lengths. Interestingly, this effect appears to amplify with increasing sequence length (loss differences across the sequence lengths: 0.0028 [1 min], 0.0129 [2 min], 0.0169 [4min]). So, the descent increase of the sampling interval may indeed mitigate the impact of noise on the derived features, yielding a balance between reducing noise and minimizing the loss of information content.
- Additionally, longer sequences, depending on a constant sampling interval, lead to a significant decrease of the validation loss. We will further discuss the cause of this effect in the next subsection.
- With respect to the remaining hyperparameters, no clear trends can be inferred from the table. While it appears that the RNNs tend towards deeper input-to-recurrent groups, a single LSTM cell and rather shallow recurrent-to-output groups, there are also individual deviations from these tendencies.

In conclusion, the found models (identified by *<sampling interval>_<seq. length in min.> (<rank>)*) can be ranked by their loss as follows: 2s_4min (1.) < 1s_4min (2.) < 2s_2min (3.) < 1s_2min (4.) < 2s_1min (5.) < 1s_1min (6.).

*B. Model Performance and Interpretation*

Based solely on the validation loss, the best performance in road user classification within the set boundaries was achieved based on the model relying on sequences with a sampling interval of 2s and a sequence length of 120 samples (4min). Normally, the subsequent evaluation would exclusively focus on this best-found model. However, we decided to evaluate multiple models based on the test data within this chapter: specifically, all models referring to a sampling interval of 2 seconds. This comparison is not intended to provide any specific recommendation for model selection but rather to offer insights into the impact of sequence length on the classification performance of the RNNs.

The results of the evaluation of these models are presented as Confusion matrices in Table IV. Inherently the number of test sequences used to build these matrices shrinks with increasing sequence length. It should be noted that this may promote random effects, negatively influencing the validity of the derived performance metrics. However, the number of sequences should be sufficient for a rough assessment of the models.

TABLE IV
CONFUSION MATRICES OF THE 2S MODELS

| | | Seq. Length: 30 (1 min) | | | |
|---|---|---|---|---|---|
| | | ped. | cyc. | mot. | car |
| True Label | ped. | **92** | 0 | 0 | 1 |
| | cyc. | 2 | **106** | 0 | 7 |
| | mot. | 0 | 1 | **62** | 32 |
| | car | 0 | 2 | 42 | **109** |

| | | Seq. Length: 60 (2 min) | | | |
|---|---|---|---|---|---|
| | | ped. | cyc. | mot. | car |
| True Label | ped. | **45** | 0 | 0 | 0 |
| | cyc. | 0 | **53** | 0 | 3 |
| | mot. | 0 | 0 | **33** | 13 |
| | car | 0 | 3 | 8 | **60** |

| | | Seq. Length: 120 (4 min) | | | |
|---|---|---|---|---|---|
| | | ped. | cyc. | mot. | car |
| True Label | ped. | **20** | 1 | 0 | 0 |
| | cyc. | 0 | **28** | 0 | 0 |
| | mot. | 0 | 0 | **15** | 6 |
| | car | 0 | 0 | 3 | **26** |

We use these confusion matrices to calculate the F1-Scores per class and the overall Macro-F1-Scores for each model and present them in Table V.

Unsurprisingly, all models demonstrate excellent performance in classifying *pedestrian* trajectories, consistently achieving high F1-Scores, with a perfect score of 1.0 for the *2 min model*. The 4min model exceptionally achieves the lowest score for pedestrians, due to a single misclassification with the *cyclist* class, which has a significant impact due to the small size of the test dataset.

TABLE V
F1-SCORES PER CLASS OF THE 2S MODELS

| | Seq. Length: 30 (1 min) | Seq. Length: 60 (2 min) | Seq. Length: 120 (4 min) |
|---|---|---|---|
| $F1_{ped.}$ | 0.9840 | 1.0000 | 0.9756 |
| $F1_{cyc.}$ | 0.9464 | 0.9464 | 0.9825 |
| $F1_{mot.}$ | 0.6231 | 0.7586 | 0.7692 |
| $F1_{car}$ | 0.7219 | 0.8163 | 0.8525 |
| **Macro-F1** | **0.8189** | **0.8803** | **0.8950** |

The models retain a high performance for the cyclist class, particularly excelling for the model using the 4 min sequence length, achieving an F1-Score of 0.9825. However, the positive correlation between the classification performance of the models and the length of the underlying sequences becomes unambiguous when considering the classes of *motorcyclists* and *passenger cars*. The consistent improvement in the capability of discriminating trajectories generated by motorized vehicles results in a F1-score of 0.7692 for motorcyclists and 0.8525 for passenger cars. This correlation is also reflected by the consistent increase of the Macro-F1-Score, which especially indicates a robust performance across all classes for the *2 min* and *4 min models*.

To give even further insights into the effect of the sequence length on the classification performance with respect to the specific classes, Fig. 2 plots the ratio of incorrectly classified test sequences against the evaluated sequence length (in timesteps) independently for each of the three models. The curves can be interpreted as representing the error rates of the models for individual classes at a specific point in time during the observation period.

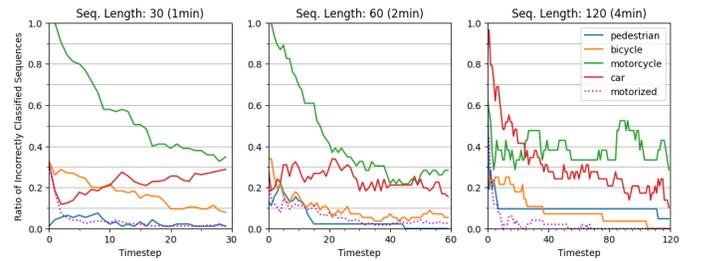

**Fig. 2.** Ratio of incorrectly classified sequences at a specific timestep for each 2s model

Note that all models seem to demonstrate the capability to give a relatively robust pre-categorization into *pedestrians*, *cyclists*, and *motorized vehicles* (*motorcycles* and *passenger cars* combined) within a temporal span of just 1 to 5 timesteps. We believe that the models likely base this pre-categorization on the leading velocity-values in the sequences, which, in many scenarios, should be sufficient to make a reasonable preliminary distinction among the three categories. (e.g., a velocity value of 50 km/h is implausible for a pedestrian, relatively unlikely for a cyclist, but typical for motorized vehicles). Among the motorized categories, either *passenger cars* or motorcyclists appear to serve as the default assumption. This becomes apparent as one of the two classes in all three models initially starts with an error rate close to 1.0, while the error rate of the other class is relatively low right from the beginning.

With increasing timesteps, the prediction of all models is inherently becoming more precise. However, with respect to the 1 min and 2 min models the underlying sequence length only seems sufficient for the *pedestrian* and arguably the *cyclist* class to reach some form of a plateau. Regarding the *motorcyclists* and *passenger car* classes, there appears to be an ongoing downward trend for at least one of both classes across all models, implying that longer sequences would likely result in further improvement of the classification performance. This assumption is supported by our experimental results, as an increase in sequence length (from 1 min to 2 min and 2 min to 4 min) obviously led to better classification performance. It can be assumed that further increase (beyond 4 min) would result in additional, albeit diminishing, improvement.

We posit that a precise classification of road users requires sequences of a certain length, because road users appear challenging to differentiate (solely on GNSS data) within many common traffic situations, e.g., think of road users waiting at traffic lights, stuck in traffic or slowed down by preceding cyclists. Distinguishing motorized vehicles, which often display similar behaviors both in specific situations and generally, appears to require longer sequences to increase the likelihood of capturing scenarios, such as acceleration or cornering, that improve predictive accuracy. As indicated by the late reductions of the classification errors in the last chart, such situations may occur late (or do not occur at all) within individual sequences.

## IV. Conclusion und future work

In response to the limitations of traditional road user classification approaches, i.e. approaches that utilize expensive and potentially intrusive fixed-point or mobile sensor systems with low spatial coverage, this work investigated an unintrusive and cost-effective alternative based on LSTM-RNNs applied to high frequency GNSS sequences, collected from general-purpose edge devices such as smartphones. As a starting point, we restricted the classification task to scenarios within the boundaries of medium-sized towns and rural communities, and to four types of road users, namely: *pedestrians*, *cyclists*, *motorcycles*, and *passenger cars*. The data specifically collected for this study consists of 165 trajectories, amounting to a total duration of approximately 32 hours.

The results of the work indicate that our method, which utilized RNNs with LSTM-cells and input-to-recurrent- as well as recurrent-to-output-feed-forward-layers, is capable of reliably classifying *pedestrians*, *cyclists*, and *motorized vehicles*, based on even short sequences. The more granular classification of motorized vehicles into *motorcyclists* and *passenger cars* is considerably more challenging. However, the capability to efficiently classify corresponding sequences has shown to improve with increasing sequence lengths. Our method achieves a F1-score of 0.82 for passenger cars and 0.76 for motorcyclists based on sequences of just two-minute length or of 0.85 and 0.77 based on four-minute sequences respectively. By classifying road users not only reliably but also based on a considerably lower duration of observation it therefore demonstrates its potential to outperform existing approaches, that utilize lower-frequency GNSS data. However, the scope of the examined classification task is currently quite limited. Future research should explore the impact of a wider range of road conditions – such as urban, rural, and highway settings – on classification performance, as different environments may affect GNSS data characteristics and influence road user behavior. Additionally, expanding the range of road user types to include more vehicle classes, such as buses and trucks, could help validate the model's generalizability across a broader set of real-world applications.

While our method offers an effective alternative, it is currently less granular and accurate than state-of-the-art fixed-point and mobile sensor approaches. Since our experiments have been carried out on a considerably small dataset, this may change with more high-frequency GNSS trajectories, featuring more traffic scenarios, becoming available to train the models. Moreover, where applicable, integrating data from other sensors, such as radars and cameras, could further enhance the accuracy and robustness of classification systems utilizing the proposed method. This multi-sensor approach could allow better differentiation, especially between vehicles, and improve performance in complex traffic environments where GNSS data alone may be limited. Furthermore, we hope that 5G positioning for street scenarios will be available in the near future such that we can pursue our original intention.

An important consideration not yet addressed in this study is the dynamic nature of road users, who may switch their mode of transportation - and consequently their *classification* - at any time in real-world traffic, like leaving a car to walk or climbing on a bicycle to get along faster. However, since our networks are designed for sequence-to-sequence classification, the dynamic classification of the transportation mode can be natively achieved with our models by just training them with correspondingly labeled data.

Beyond that, future research may apply and extent similar approaches to other related problems, such as driving behavior identification or determining the relevance of road users in an extended environment with respect to a specific autonomous *hero*-vehicle. Moreover, efforts should include the evaluation of alternative network architectures to further enhance the robustness and performance of the models. This may involve investigating the use of other recurrent cells, such as *Gated Recurrent Units* (GRU-cells), *attention mechanisms* [27], or *one-dimensional convolutional layers*.

Finally, it is crucial to address some ethical, legal and privacy concerns associated with our method. Since adversaries can use a similar approach to discover hidden information from road users and use it to re-identify a target, advanced privacy methods may need to be developed to resolve such issues.

48IoT-41373-2024

Methods and Software Perspective," *IEEE Open J. Intell. Transp. Syst.*, vol. 2, pp. 173–194, 2021, doi: 10.1109/OJITS.2021.3096756.

[5] M. Simoncini, L. Taccari, F. Sambo, L. Bravi, S. Salti, and A. Lori, "Vehicle classification from low-frequency GPS data with recurrent neural networks," *Transp. Res. Part C Emerg. Technol.*, vol. 91, pp. 176–191, Jan. 2018, doi: 10.1016/j.trc.2018.03.024.

[6] Z. Sun and X. (Jeff) Ban, "Vehicle classification using GPS data," *Transp. Res. Part C Emerg. Technol.*, vol. 37, pp. 102–117, Dec. 2013, doi: 10.1016/j.trc.2013.09.015.

[7] S. H. Tan, J. H. Chuah, C.-O. Chow, J. Kanesan, and H. Y. Leong, "Artificial intelligent systems for vehicle classification: A survey," *Eng. Appl. Artif. Intell.*, vol. 129, no. C. Pergamon Press, Inc., p. 20, 2024.

[8] H. Shokravi, H. Shokravi, N. Bakhary, M. Heidarrezaei, S. S. Rahimian Koloor, and M. Petrů, "A Review on Vehicle Classification and Potential Use of Smart Vehicle-Assisted Techniques," *Sensors*, vol. 20, no. 11, p. 3274, Jun. 2020, doi: 10.3390/s20113274.

[9] Z. Sun and X. Ban, "Vehicle classification using GPS data," *Transp. Res. Part C Emerg. Technol.*, vol. 37, pp. 102–117, Jan. 2013, doi: 10.1016/j.trc.2013.09.015.

[10] M. Kafai and B. Bhanu, "Dynamic Bayesian Networks for Vehicle Classification in Video," *IEEE Trans. Ind. Inform.*, vol. 8, no. 1, pp. 100–109, Feb. 2012, doi: 10.1109/TII.2011.2173203.

[11] J. Barreyro, L. R. Yoshioka, C. L. Marte, C. G. Piccirillo, M. M. D. Santos, and J. F. Justo, "Assessment of Vehicle Category Classification Method Based on Optical Curtains and Convolutional Neural Networks," *IEEE Access*, vol. 12, pp. 133532–133544, 2024, doi: 10.1109/ACCESS.2024.3410160.

[12] S. Dabiri, N. Markovic, K. Heaslip, and C. K. Reddy, "A deep convolutional neural network based approach for vehicle classification using large-scale GPS trajectory data," *Transp. Res. Part C*, vol. 116, 2020, doi: https://doi.org/10.1016/j.trc.2020.102644.

[13] Y. Zhou, H. Nejati, T.-T. Do, N.-M. Cheung, and L. Cheah, "Image-based Vehicle Analysis using Deep Neural Network: A Systematic Study," Aug. 07, 2016, *arXiv*: arXiv:1601.01145. Accessed: Oct. 04, 2022. [Online]. Available: http://arxiv.org/abs/1601.01145

[14] T. Gebru, J. Krause, Y. Wang, D. Chen, J. Deng, and L. Fei-Fei, "Fine-Grained Car Detection for Visual Census Estimation," *Proc. AAAI Conf. Artif. Intell.*, vol. 31, no. 1, Feb. 2017, doi: 10.1609/aaai.v31i1.11174.

[15] S. Dwivedi *et al.*, "Positioning in 5G Networks," *IEEE Commun. Mag.*, vol. 59, no. 11, pp. 38–44, Nov. 2021, doi: 10.1109/MCOM.011.2100091.

[16] F. Mogyorósi *et al.*, "Positioning in 5G and 6G Networks—A Survey," *Sensors*, vol. 22, no. 13, p. 4757, Jun. 2022, doi: 10.3390/s22134757.

[17] S. Stephenson, "Automotive applications of high precision GNSS," Thesis, University of Nottingham, Nottingham, 2016. Accessed: Dec. 07, 2023. [Online]. Available: https://eprints.nottingham.ac.uk/38716/1/Scott%20Stephenson%20-%204148128%20-%20Thesis%2015%2011%202016.pdf

[18] S. Stephenson, X. Meng, T. Moore, A. Baxendale, and T. Edwards, *Accuracy Requirements and Benchmarking Position Solutions for Intelligent Transportation Location Based Services*. 2011.

[15] S. Dabiri and K. Heaslip, "Transport-domain applications of widely used data sources in the smart transportation: A survey," *CoRR, abs/1803.10902*, 2018.

[20] X. Song, H. Kanasugi, and R. Shibasaki, "Deeptransport: Prediction and simulation of human mobility and transportation mode at a citywide level," in *Proceedings of the Twenty-Fifth International Joint Conference on Artificial Intelligence (IJCAI-16)*, New York, NY, USA, 2016, pp. 2618–2624.

[21] S. Dabiri and K. Heaslip, "Inferring transportation modes from GPS trajectories using a convolutional neural network," *Transp. Res. Part C Emerg. Technol.*, vol. 86, pp. 360–371, Jan. 2018, doi: 10.1016/j.trc.2017.11.021.

[22] R. Zhang, P. Xie, C. Wang, G. Liu, and S. Wan, "Classifying transportation mode and speed from trajectory data via deep multi-scale learning," *Comput. Netw.*, vol. 162, p. 106861, 2019.

[23] J. Trivedi, M. S. Devi, and B. Solanki, "Step Towards Intelligent Transportation System With Vehicle Classification and Recognition Using Speeded-Up Robust Features," *Arch. Tech. Sci.*, vol. 1, no. 28, pp. 39–56, Jun. 2023, doi: 10.59456/afts.2023.1528.039J.

[24] S. Tas, O. Sari, Y. Dalveren, S. Pazar, A. Kara, and M. Derawi, "Deep Learning-Based Vehicle Classification for Low Quality Images," *Sensors*, vol. 22, no. 13, p. 4740, Jun. 2022, doi: 10.3390/s22134740.

[25] A. Sherstinsky, "Fundamentals of Recurrent Neural Network (RNN) and Long Short-Term Memory (LSTM) Network," *Phys. Nonlinear Phenom.*, vol. 404, p. 132306, Mar. 2020, doi: 10.1016/j.physd.2019.132306.

[26] H. Salehinejad, S. Sankar, J. Barfett, E. Colak, and S. Valaee, "Recent Advances in Recurrent Neural Networks," Feb. 22, 2018, *arXiv*: arXiv:1801.01078. Accessed: Nov. 15, 2024. [Online]. Available: http://arxiv.org/abs/1801.01078

[27] A. Vaswani *et al.*, "Attention Is All You Need," presented at the 31st Conference on Neural Information Processing System, Long Beach, CA, USA: arXiv, 2017. doi: 10.48550/ARXIV.1706.03762.

[28] J. B. Tenenbaum and W. T. Freeman, "Separating Style and Content with Bilinear Models," *Neural Comput.*, vol. 12, no. 6, pp. 1247–1283, Jun. 2000, doi: 10.1162/089976600300015349.

[29] D. P. Kingma and J. Ba, "Adam: A method for stochastic optimization," *ArXiv Prepr. ArXiv14126980*, 2014.

[30] T. Mikolov, M. Karafiát, L. Burget, J. Cernocký, and S. Khudanpur, "Recurrent neural network based language model.," in *Interspeech*, Makuhari, 2010, pp. 1045–1048.

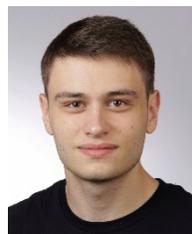

**Lennart Köpper** is a graduate data science student at Coburg University of Applied Sciences and Arts. There he previously received his bachelor's degree in computer science. He has practical experience in software engineering, application development, test automation and machine learning.

IoT-41373-2024

Initially as a student assistant and later as a laboratory engineer, he contributed to the research project '5G-KC' focused on enhancing the situational awareness of autonomous vehicles by expanding their perceptional capabilities using shared mobile communications data, distributed edge computing and machine learning. His main (research) interests include data mining, deep learning and data engineering.

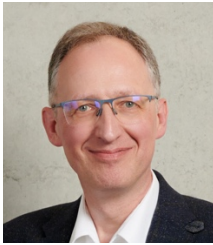

**Thomas Wieland** is full professor of computer science at Coburg University of Applied Sciences and Arts. His main research interests are Internet of Things, environmental monitoring, wireless sensor networks, and machine learning applications. Until 2020 he headed the Fraunhofer Application Center for Wireless Sensor Systems, part of Fraunhofer Institute for Integrated Circuits IIS.

He focuses on the entire data processing chain from sensors and other active data sources up to their incorporation in so-called real-world applications. He applies these technologies to the domains of smart cities, smart factories, pervasive computing, intelligent transportation systems, and agriculture. He is a member of various program committees and organizing committees of respective conferences and workshops (e.g. Open Identity Summit, Wirtschaftsinformatik).